\documentclass[lettersize,journal]{IEEEtran}
\usepackage{amsmath,amsfonts}
\usepackage{algorithmic}
\usepackage{array}
\usepackage[caption=false,font=normalsize,labelfont=sf,textfont=sf]{subfig}
\usepackage{textcomp}
\usepackage{stfloats}
\usepackage{url}
\usepackage{verbatim}
\usepackage{graphicx}
\def\BibTeX{{\rm B\kern-.05em{\sc i\kern-.025em b}\kern-.08em
    T\kern-.1667em\lower.7ex\hbox{E}\kern-.125emX}}
\usepackage{balance}
\usepackage{booktabs}
\usepackage{multirow}
\usepackage[numbers]{natbib}
\usepackage{makecell}
\begin{document}
\title{FLaTEC: Frequency-Disentangled Latent Triplanes for Efficient Compression of LiDAR Point Clouds}
\author{Xiaoge Zhang, Zijie Wu, Mingtao Feng, Zichen Geng, Mehwish Nasim, Saeed Anwar, Ajmal Mian
\thanks{Xiaoge Zhang, Zijie Wu, Zichen Geng, Mehwish Nasim, Saeed Anwar, and Ajmal Mian are with the University of Western Australia, Crawley, Perth, WA 6009, Australia (Email: xiaoge.zhang@research.uwa.edu.au; zen.geng@research.uwa.edu.au; wuzijieeee@hnu.edu.cn; mehwish.nasim@uwa.edu.au; saeed.anwar@uwa.edu.au; ajmal.mian@uwa.edu.au); Mingtao Feng is with Xidian University, Xi'an 710071, China (Email: mintfeng@hnu.edu.cn).}}

\maketitle

\begin{abstract}
Point cloud compression methods jointly optimize bitrates and reconstruction distortion. However, balancing compression ratio and reconstruction quality is difficult because low-frequency and high-frequency components contribute differently at the same resolution. To address this, we propose FLaTEC, a frequency-aware compression model that enables the compression of a full scan with high compression ratios. Our approach introduces a frequency-aware mechanism that decouples low-frequency structures and high-frequency textures, while hybridizing latent triplanes as a compact proxy for point cloud. Specifically, we convert voxelized embeddings into triplane representations to reduce sparsity, computational cost, and storage requirements. We then devise a frequency-disentangling technique that extracts compact low-frequency content while collecting high-frequency details across scales. The decoupled low-frequency and high-frequency components are stored in binary format. During decoding, full-spectrum signals are progressively recovered via a modulation block. Additionally, to compensate for the loss of 3D correlation, we introduce an efficient frequency-based attention mechanism that fosters local connectivity and outputs arbitrary resolution points. Our method achieves state-of-the-art rate-distortion performance and outperforms the standard codecs by 78\% and 94\% in BD-rate on both SemanticKITTI and Ford datasets.
\end{abstract}


\section{Introduction}
\IEEEPARstart{P}{oint} cloud is a fundamental 3D data representation and widely adopted in numerous applications, including autonomous driving~\cite{lin2024bev}, robotics~\cite{zhu2024point}, virtual reality~\cite{guo2020deep}, \emph{etc}. However, high-resolution LiDAR scans typically generate massive amounts of data, posing significant challenges for storage and transmission. In this paper, we propose an efficient frequency-disentangled compression framework that 
enables scalability, fast decoding, and high-fidelity point cloud compression.

\begin{figure}
    \centering
    \includegraphics[width=0.95\linewidth, trim=0cm 10.5cm 17cm 1.5cm]{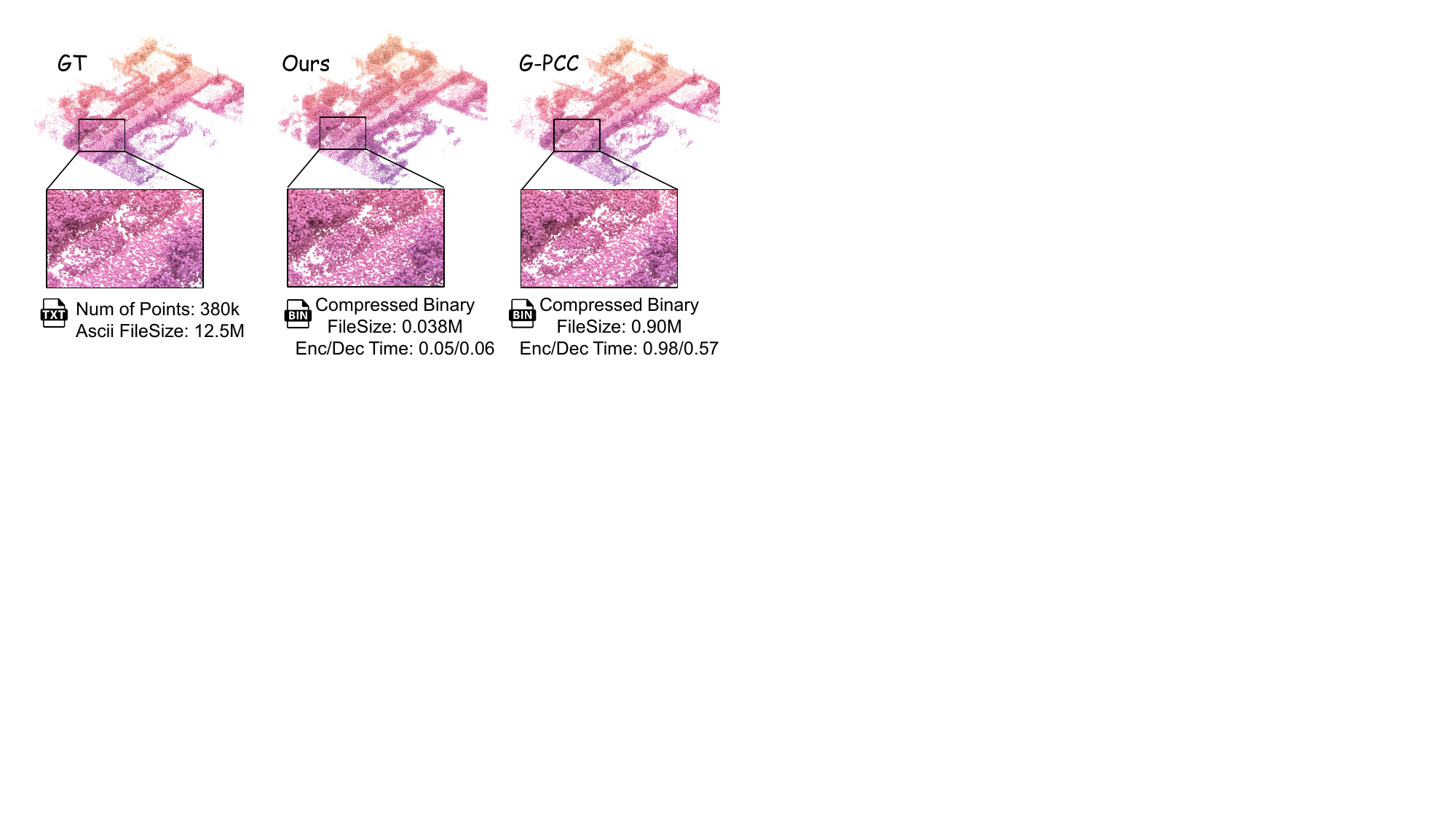}
    \vspace{-4mm}
    \caption{Qualitative and quantitative comparison of point cloud compression methods. The proposed FLaTEC substantially reduces both the compressed file size and encoding/decoding time, while achieving comparable reconstruction quality to the baseline. The zoom-in region highlights vehicles on the road. 
    }
    \vspace{-4mm}
    \label{fig:quick-run-top}
\end{figure}

In recent years, deep learning-based compression methods~\cite{quach2022survey, roriz2024survey} have been proposed to minimize consumed bitrates and reconstruction distortion simultaneously. Point-based method using specialized architectures like PointNet++\cite{qi2017pointnet++} or PointTransformer\cite{zhao2021point}, where KNN-based feature aggregation is employed to capture local geometric structures~\cite{wiesmann2021deep, he2022density, huang20223qnet}. However, these methods struggle with large-scale LiDAR point clouds due to memory usage scaling linearly with the number of points, which often exceeds 12,000 per scan. To address this, voxel-based methods~\cite{wang2021multiscale, nguyen2021multiscale} discretize 3D space into a fixed-resolution voxel grid, where each voxel indicates whether its spatial region is occupied. This transformation decouples memory usage from the number of input points and scales more effectively to large outdoor scenes. However, voxel-based methods still suffer from cubic complexity, resulting in \textit{Problem 1)}: high computational cost and inference latency. Also, the inherent sparsity of LiDAR data introduces severe voxel label imbalance, leading to training instability.
Furthermore, LiDAR-based applications demand high-fidelity decompression. Due to the absence of skip connections in compression tasks, preserving essential structural details under limited bitrate constraints remains a significant challenge.

\begin{figure}
    \centering
    \includegraphics[width=0.95\linewidth, trim=0cm 6.5cm 18.5cm 0cm]{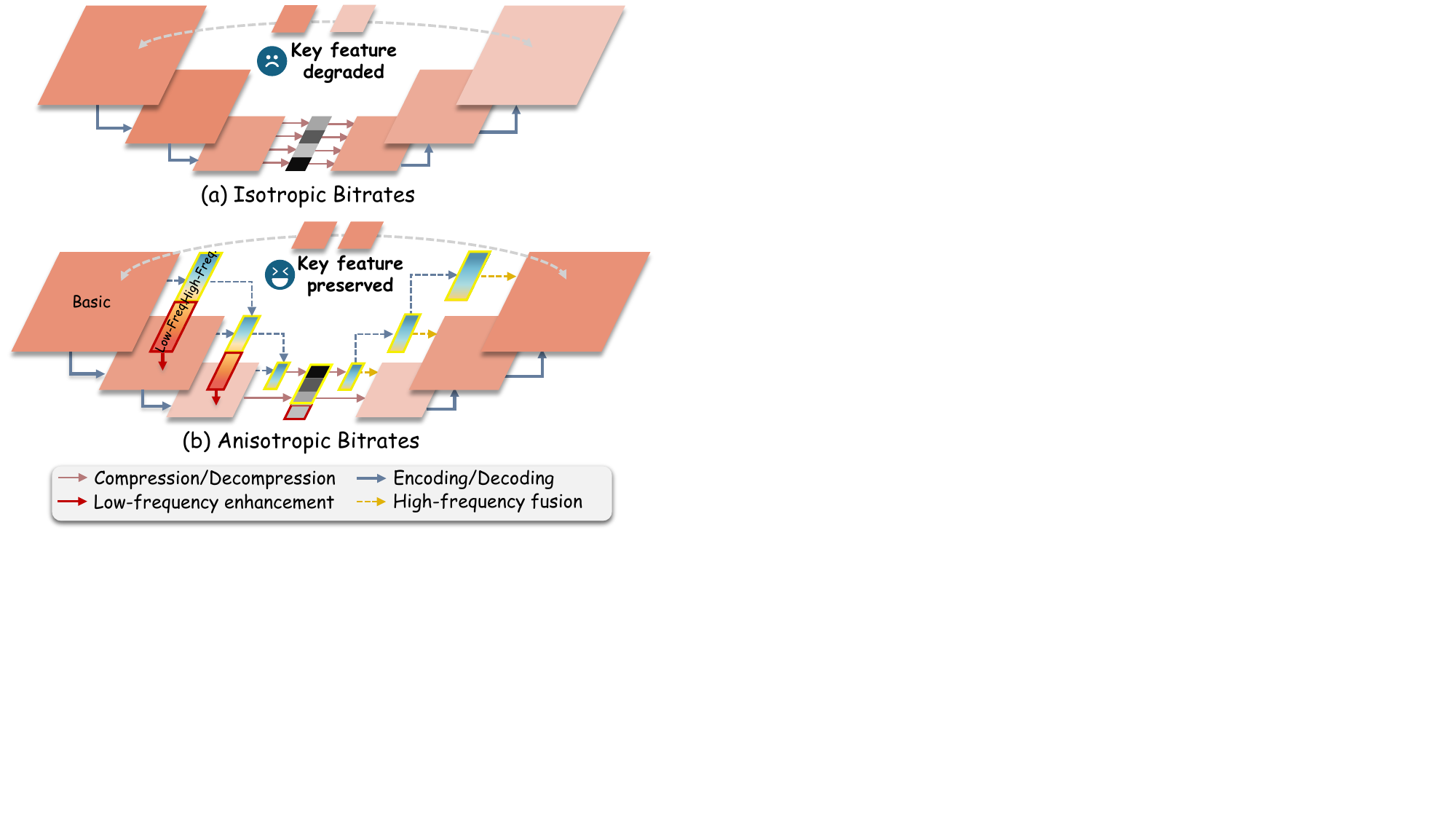}
    \vspace{-4mm}
    \caption{(a) Traditional deep learning methods encode all frequency components at the same resolution, resulting in a suboptimal trade-off for compression. (b) Our method disentangles high-frequency details from basic features, allowing for flexible bitrate allocation across different levels of detail. 
    }
    \vspace{-4mm}
    \label{fig:quick-run}
\end{figure}

When LiDAR scans are encoded in the spatial domain, high-frequency components receive insufficient bits, causing details loss and degraded reconstruction, as illustrated in Figure~\ref{fig:quick-run}. This `isotropic bitrates' problem stems from treating all frequencies equally despite their differing perceptual importance. To address this, some methods introduce separate encoding branches for high- and low-frequency components~\cite{zhou2023xnet}, while others employ multi-directional windowed attention to simulate frequency-aware
decomposition~\cite{li2023frequency}. Although promising, these techniques raise \textit{Problem 2)}: These methods produce limited and ambiguous frequency decompositions, restricting flexible spectral control over the contribution of each component to the reconstruction.

In this paper, we propose a frequency-aware triplane-based compression model tailored for large-scale LiDAR point clouds, enabling efficient representation with minimal information loss. To address the challenges of sparsity and scalability, we decompose voxelized points into three orthogonal 2D planes, drawing inspiration from PCA to capture dominant spatial variations. This projection strategy mitigates data sparsity in unoccupied regions of LiDAR point clouds, enabling denser supervision and more compact encoding. It also reduces the computational and storage complexity from cubic $O(\mathcal{H}\mathcal{W}\mathcal{D})$ to a quadratic $O(\mathcal{H}\mathcal{W} + \mathcal{H}\mathcal{D} + \mathcal{W}\mathcal{D})$, significantly improving compression efficiency.
To achieve anisotropic bitrates, we introduce a stage-wise disentangling mechanism that progressively decomposes the mixed spectral information. By boosting the core features with low-frequency components, while aggregating fine-grained details from the high-frequency spectrum to a hierarchical representation, the framework enables adaptive bitrate allocation. 
Additionally, to further refine the reconstructed 3D geometry, we resort to the frequency domain dot-product to calculate self-attention within a specific window, thereby developing regional connectivity.

Our contributions are summarized as follows:

\begin{itemize}
    \item We propose a frequency-aware method for compressing LiDAR scans at high compression ratios. 

\item We introduce a triplane-based compression model that alleviates the need for expensive 3D operations. 

\item  We present a locality-driven refinement module which employs an efficient frequency-based voxel-wise attention mechanism and a plug-and-play point upsampler to compensate for quantization errors in voxel data. Hence, LiDAR scans can be upsampled to the desired resolution with improved reconstruction accuracy.

\end{itemize}

\section{Related Work}
\vspace{1mm}\noindent\textbf{Point-based compression methods} leverage architectures such as KPConv~\cite{thomas2019kpconv}, PointNet++~\cite{qi2017pointnet++}, and PointTransformer~\cite{zhao2021point} to aggregate neighbourhood information. The reduced points, along with aggregated features, are subsequently quantized and stored as a binary file. Certain approaches~\cite{zhang2022transformer, almuzaddidVariableRateCompression2022} concentrate on compressing small-scale point clouds, while others~\cite{he2022density, huang2023patch} address the compression of large-scale point clouds through the partitioning of data into blocks or patches. This block-based partitioning results in a constrained receptive field and the loss of the complete scene context, thereby rendering these methods unsuitable for practical applications such as real-time capture and transmission. Pointsoup~\cite{you2024pointsoup} demonstrates 
 the scalability by encoding point clouds into `bone and skin' features, but suffers from a low compression ratio.

\vspace{1mm}\noindent\textbf{Octree-based compression methods} utilize an octree structure to represent point clouds, wherein each node encodes the occupancy status of eight octants, providing 256 potential occupancy labels~\cite{biswas2020muscle, fu2022octattention, que2021voxelcontext, song2023efficient}. Approaches such as OctAttention~\cite{fu2022octattention} and VoxelContextNet~\cite{que2021voxelcontext} integrate contextual information from parent and ancestor nodes, with inter-scale weights being collaboratively shared and optimized to preserve consistent correlations across varying scales. Nevertheless, these approaches suffer from slow decoding speeds due to the dependency of non-leaf nodes on their sibling nodes. The EHEM~\cite{song2023efficient} addresses this limitation by dividing nodes into two disjoint groups, thus facilitating parallel inference within each group. Despite these advancements, octree-based methodologies remain susceptible to the sparse distribution characteristic of LiDAR scans and require exhaustive octree construction and point cloud reconstruction.

\vspace{1mm}\noindent\textbf{Voxel-based compression methods} ~\cite{you2025reno, wang2022sparse} involve voxelization and partitioning of the input point cloud, utilizing three-dimensional convolutional neural networks (CNNs) for occupancy prediction. In lossless compression, VoxelDNN~\cite{nguyenLearningBasedLosslessCompression2021} and PCGC~\cite{wang2019learned} leverage a CNN backbone, but suffer from slow coding times. MSVoxelDNN~\cite{nguyen2021multiscale} and PCGCv2~\cite{wang2021multiscale} enhance their methodologies by partitioning each voxel into eight progressive groups for inter-scale occupancy prediction, thereby reducing coding latency. Nevertheless, they continue to encounter challenges in scaling to large-scale sparse LiDAR scans. Lossy compression techniques~\cite{wang2022sparse} threshold predicted occupancy probabilities to facilitate voxel reconstruction, thus enabling the handling of large outdoor point clouds, although they still face difficulties related to the imbalance between occupied and unoccupied voxels. To address sparsity in 3D space, we project the point cloud onto three orthogonal views, generating a sparse proxy of voxel features. The triplane representation reduces computational overhead while preserving the same receptive field, in comparison to sparse 3D convolutions.

\begin{figure*}[tbp]
    \centering
    \includegraphics[width=\linewidth, trim=0cm 10cm 0cm 0cm]{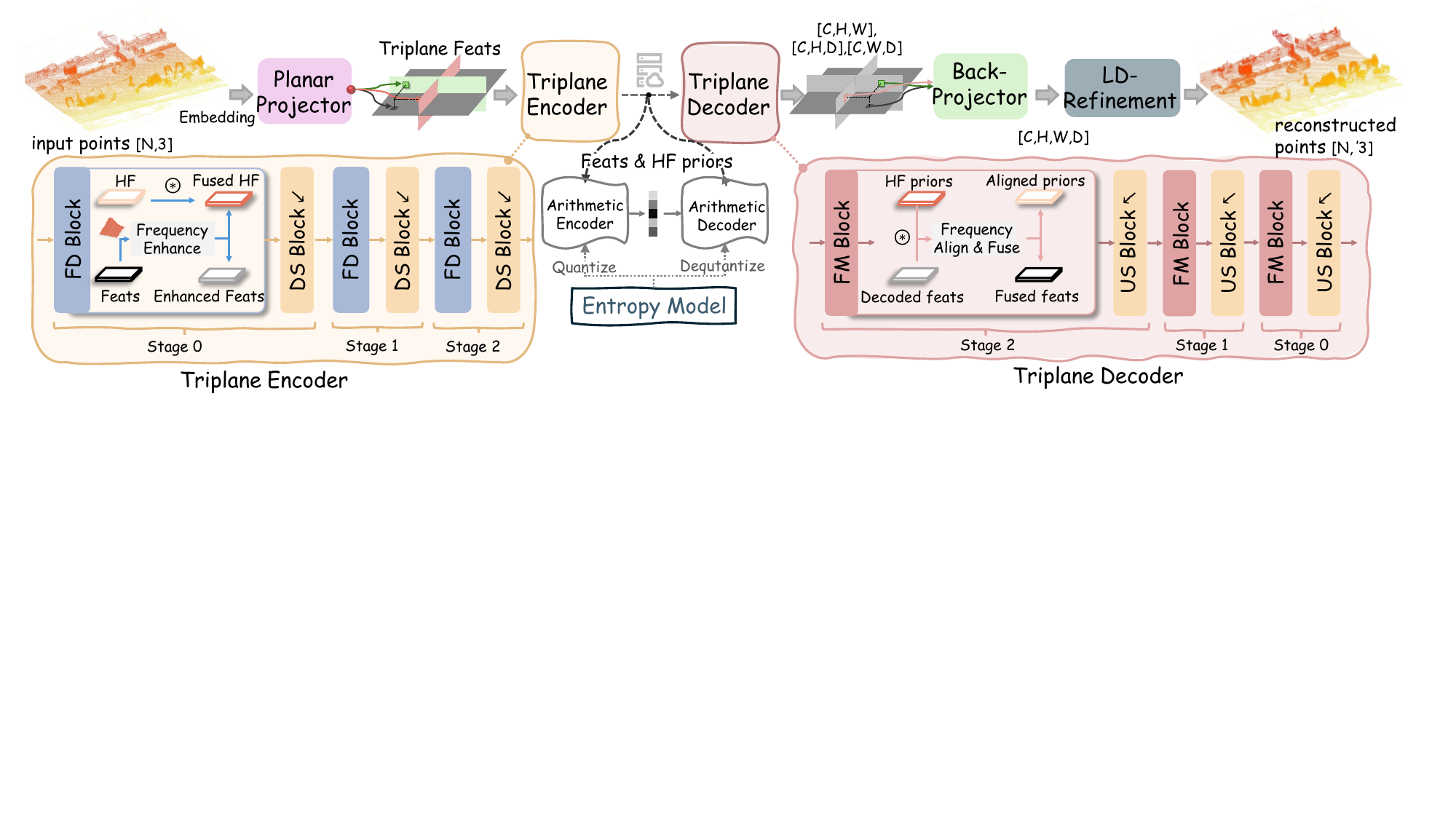}
    \vspace{-4mm}
    \caption{Overview of our compression method. FD and FM refer to feature decomposition and frequency modulation. LSA represents local spectrum attention. HF is high frequency. Voxel features are initially projected onto three orthogonal views—top, front, and side—to reduce sparsity and storage costs. These projected triplane features are then processed through separate 2D encoders, which output global content and high-frequency priors. The encoded features are subsequently quantized and converted into a binary string. During decoding, 2D decoders reconstruct fine-grained triplane features guided by high-frequency priors. Finally, the voxel features are refined with spatial correlations before generating volumetric occupancy probability.}
    \vspace{-4mm}
    \label{figs:ppl}
\end{figure*}

\paragraph{Frequency-Enhanced Feature Learning.} Wala~\cite{zhou2023xnet} and Make-a-Shape~\cite{hui2024make} employ wavelet coefficients as latent features for shape generation, attaining a substantial reduction in the coefficients. A transformer model for image deblurring~\cite{kong2023efficient} is developed, wherein both the feed-forward network and local attention are computed within the frequency domain. X-Net~\cite{zhou2023xnet} introduces dual branches to separately encode high-frequency and low-frequency features, which are subsequently fused within the decoding head for medical image segmentation. However, these approaches are not directly applicable to compression, as skip connections are not permissible in compression models, as the decoder cannot access earlier encoded features, and the only available information pertains to the quantized features in the entropy bottleneck. In image compression, In \cite{li2023frequency}, attention blocks with varying window sizes are employed to extract spectral components. Nevertheless, this technique results in a discrete decomposition confined to a limited and ambiguous frequency domain. We propose a stage-wise frequency decomposition and integration mechanism, facilitating a continuous and comprehensive spectral representation.

\section{Methodology}
\label{sec:method}
\subsection{Triplane Representation}
Given a voxelized point cloud $V\in{\mathbb{Z}^{\mathcal{H}\times\mathcal{W}\times\mathcal{D}}}$, 
the goal of our FLaTEC framework is to transform the input voxels into compressed representations and, based on that, predict the occupancy probability of each 3D voxel. Traditional voxel-based methods incur cubic complexity 
$O(\mathcal{H} \times \mathcal{W} \times \mathcal{D})$ in both computation and storage. What’s worse, in real-world outdoor point clouds~\cite{behley2019semantickitti}, a large portion of the scene is empty. While existing voxel-based methods use sparse convolutions to reduce computational load, training on large-scale scene-level 3D volumes remains challenging. To address this, we employ latent triplanes as intermediate representations for efficient LiDAR compression, as shown in Figure~\ref{figs:ppl} (see supplementary for detailed architecture). 

\vspace{1mm}\noindent\textbf{Plane Projection.} 
We first learn geometric embeddings $F_{V}\in{R^{\mathcal{H}\times\mathcal{W}\times\mathcal{D}\times\mathcal{C}_1}}$ 
from the occupancy volume, which effectively describes the local variations along the 3D space. We then perform vector-wise feature aggregation by pooling over fixed-size groups. The features within each group are then aggregated along the corresponding spatial axis via an MLP. For example, the voxel feature projection to the top view can be expressed as: 
\begin{equation}
    X_{\mathcal{H}\mathcal{W}} = (\Phi_1(\Re_1(\mathcal{P}(F_V)))),
\end{equation}
\noindent where $X_{\mathcal{H}\times\mathcal{W}}\in{\mathbb{R}^{\mathcal{H}\times\mathcal{W}\times\mathcal{C}_2}}$, $\Phi$, $\Re$ and $\mathcal{P}$ denote the MLP, rearrangement and 1D group pooling operators, respectively. The operation squeezes the 3D feature volumes $F_{V}$ into a compact planar representation. By leveraging lightweight triplane encoders, 2D features can be further compressed, resulting in substantially reduced storage requirements and model size.

\vspace{1mm}\noindent\textbf{Planar Back-Projection.} To recover the triplanar representation $\tilde{X}_{\mathcal{H}\mathcal{W}}$ into volumetric features, we apply a sub-voxel MLP followed by axis-wise replication for $N_g$ steps. The back-projection operation ($BP$) is defined as 
    $BP(X) = \mathcal{T}(\Re_2(\Phi_2((X)))$, 
where $\mathcal{T}$ denotes tiling operation. The feature volume is recovered by summing the three back-projected volumes along each axis:
$\tilde{F}_V = BP(\tilde{X}_{\mathcal{H}\mathcal{W}})+BP(\tilde{X}_{\mathcal{H}\mathcal{D}})+BP(\tilde{X}_{\mathcal{W}\mathcal{D}})$.

The sinusoidal positional embeddings are stacked to raise the awareness of spatial distribution, given the coarse prediction $F_{V}^{\text{init}}=\mathcal{CB}_{3D}(\tilde{F}_V\Vert{PE}(V))$, where $\mathcal{CB}_{3D}$ denotes a group of 3D convolutional blocks, $\Vert$ denotes concatenation.

\subsection{Spectrum-driven Triplane Analysis and Synthesis Transform} 
Recent studies by~\cite{li2023frequency} have demonstrated that applying attention mechanisms with varying window sizes allows the model to focus on different frequency components. However, this approach still results in a discrete decomposition of latent representations within a limited and ambiguous frequency domain. To facilitate a comprehensive and continuous spectral analysis of triplane features tailored for compression, we propose a stage-wise frequency decomposition and integration framework. As depicted in Figure~\ref{fig:quick-run}, our method decomposes continuous frequency components and selectively allocates bitrate budgets, thereby preserving critical features while attaining a high compression ratio. For simplicity, we designate the top view $X_{\mathcal{H}\mathcal{W}}$, with the front and side views undergoing identical operations.

\begin{figure}[tbp]
    \centering
    \includegraphics[width=\linewidth, trim=1cm 3cm 15.5cm 0cm]{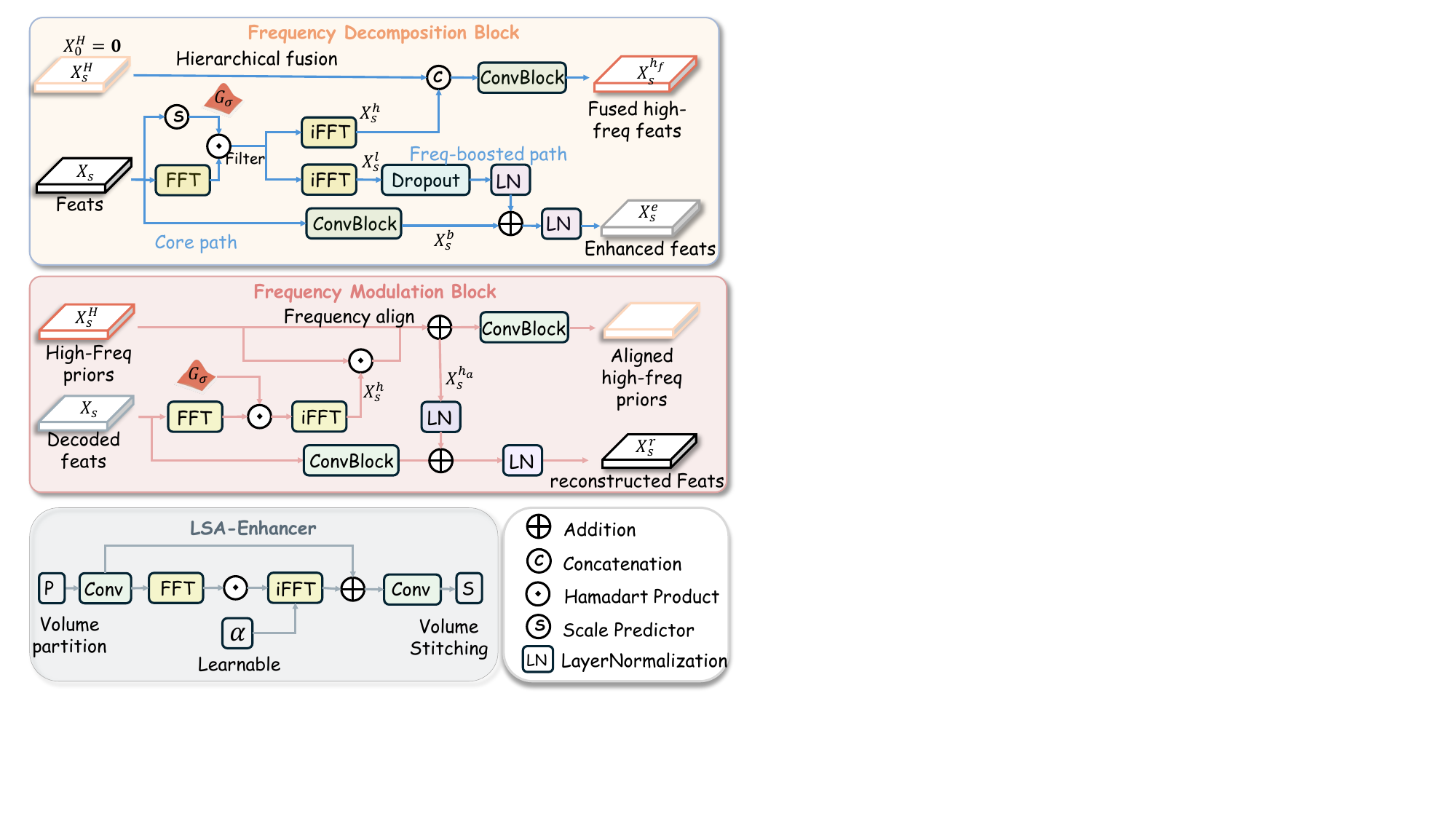}
    \vspace{-4mm}
    \caption{Module architectures. 1) The FD Block first performs frequency decomposition, then enhances global content with low-frequency structures and organizes high-frequency details into a hierarchical representation. 2) The FM Block aligns the high-frequency priors with the current-level base features, then reconstructs original details guided by the aligned priors. 3) The LSA-Enhancer refines local volumetric textures by modulating regional frequency components using a learnable attention mechanism.
    }
    \vspace{-4mm}
    \label{fig:block-freq}
\end{figure}

\vspace{1mm}\noindent\textbf{Kernel-aware Frequency-disentangled Encoder.} 
As shown in Figure~\ref{figs:ppl}, our triplane encoder is composed of alternating frequency decomposition (FD) and downsampling (DS) blocks. The FD blocks separate planar features into compact low-frequency and detailed high-frequency components, while the DS blocks progressively reduce spatial resolution. The encoder takes the extracted triplane features as its base input and leverages high-frequency (HF) priors as auxiliary input. Since the high-frequency (HF) priors are absent in the first block, we initialize them to zero. At each subsequent stage, the blocks process the base features along with the accumulated high-frequency priors from the previous stage. This process can be formally represented as:
\begin{equation}
    (X^{\mathrm{H}}_{s+1}, X_{s+1}) = \mathcal{DS} (\mathcal{FD}(X^{\mathrm{H}}_{s}, X_{s} )), s = 0, 1, \dots, S
\end{equation}with $X_0 = X$, $X^{\mathrm{h}}_0 = \mathbf{0}$.

\noindent\textit{1) Frequency Decomposition Block.}
As shown in Figure~\ref{fig:block-freq}, the encoding of basic features consists of two paths: the core path and the frequency-boosted path. The core path maintains essential features, whereas the frequency-boosted path utilizes low-frequency components to improve global awareness of base features. High-frequency components are aggregated progressively over multiple stages.

At stage $s$, the core path convolves with input $X_s$ to get core features $X_s^{c}$. On the other hand, the frequency-boosted path applies a Discrete Fourier Transform (DFT) and a frequency shift to $X_s$. This results in frequency components that follow an isotropic distribution, where low frequencies are in the center, and high frequencies are positioned at the periphery of the spectrum map. 

To decouple the frequency spectrum, we apply a soft mask to obtain low-frequency information parameterized by a Gaussian Filter:
\begin{equation}
    X_{s}^{l}=\mathcal{F}^{-1}(\mathcal{F}(X_{s})\odot(\theta(X_s)\cdot{G_{\sigma}})),
\end{equation}
where $\mathcal{F}$ and $\mathcal{F}^{-1}$ denote DFT and its inverse, respectively. $G_{\sigma}$ is a Gaussian filter with a standard deviation $\sigma$, whose magnitude is dynamically controlled based on the spatial characteristics of input features by the scale predictor $\theta$. After the filtering process, the global geometry is extracted from the central region of the frequency map, while the high-frequency components correspond to the remaining terms.

Next, to amplify the core features with general texture and overall structure, the core features are augmented by incorporating the distilled low-frequency components $ 
    X_{s}^{e}=\sigma(X_s^{c}+\sigma(\dagger(X_{s}^{l}))),$
where $\sigma$ and $\dagger$ represent layer normalization and dropout, respectively.
Similarly, the high-frequency texture is fused to accumulate high-frequency information across multiple stages by
    $X_{s}^{h_f}=\mathcal{CB}(X_s^{h}\Vert\mathcal{CB}(X_{s}^{H}))$.

\noindent\textit{2) Downsample Block.}
The enhanced features $X_{s}^{e}$ and fused high-frequency components $X_{s}^{h_f}$ are upsampled through a strided convolution backbone and forwarded to produce $X_{s+1}$ and $X_{s+1}^{H}$. This iterative process continues until the final features are encoded as binary string for transmission and storage. To minimize overall compression bitrate, we jointly optimize the likelihoods of augmented base features and high-frequency components, as detailed in Section~\ref{sec:loss}, enabling flexible bitrate allocation between them. 

\vspace{1mm}\noindent\textbf{Level-aligned Frequency Modulation Decoder.}
The triplane decoder consists of interleaved frequency modulation (FM) and upsampling (US) blocks, progressively recovering planar features before passing them to the back-projector for voxel occupancy prediction. The FM blocks leverage stored high-frequency priors for multi-level high-frequency reconstruction, while the US blocks gradually recover the triplane features to the desired spatial resolution. 

\noindent\textit{1) Frequency Modulation Block.}
The input to the decoder comprises dequantized features derived from the binary bitstring, divided into two parts: the decoded global content $X_S$ and the fully fused high-frequency details $X_S^H$. The high-frequency components cannot be used directly due to scale misalignment between the decoded base features and the high-frequency details. 

To enforce spectrum-wise consistency at a specific stage $s$, we first align the high-frequency spectrum $\tilde{X}_s^H$ with the underlying features $\tilde{X}_s$:
$\tilde{X}_{s}^{h_a}=\tilde{X}_s^H+\tilde{X}_s^H\odot\tilde{X}_s^{h}$.
Note that $\tilde{X}_s^{h}=\mathcal{F}^{-1}(\mathcal{F}(\tilde{X}_s)\odot(G(\sigma)))$. The stage order is reversed during decoding for clarity, and \( s = S, S-1, \dots, 0 \) indicates stages from the coarsest to the finest level. After aligning the high-frequency components, they are used to guide the decoding of reconstructed features $X_{S}$: $
    \tilde{X}_s^{r}=\operatorname{LN}(\mathcal{CB}(\tilde{X}_s)+\operatorname{LN}(\tilde{X}_s^{h_a}))$.

\noindent\textit{2) Upsample Block.}
Mirroring the DS block, each US block restores the spatial resolution of the reconstructed global features and aligned high-frequency priors via a transposed convolutional backbone, yielding outputs $X_{s+1}$ and $X_{s+1}^{H}$.

\subsection{3D Locality-Driven Refinement}
Compared to the original point cloud, the coarsely reconstructed volume lacks adequate 3D structural cues and suffers from quantization errors due to voxelization. Hence, we propose a two-stage refinement framework for improved accuracy: first, voxel-wise local connectivity is considered through computationally efficient frequency-domain operations; second, point cloud density is adaptively increased to arbitrary resolutions with a plug-and-play upsampler.

\vspace{1mm}\noindent\textbf{Local Spectrum-attention Enhancer.}

Our goal is to enhance local connectivity within the 3D neighborhood via attention mechanisms, which often incur high computational costs. By leveraging direct access to the frequency spectrum and applying the convolution theorem,

we safely replace the traditional convolution kernel, which is convolved with the entire image, with adaptive learnable parameters that utilize dot-product operations to obtain windowed saliency. The Specturm-attention is defined as:
\begin{equation}
    \operatorname{SA}=\mathcal{F}^{-1}(\alpha\odot\mathcal{F}(F_V))+F_V.
\end{equation}
Here, $\alpha$ is learned across all samples. The volume is partitioned into blocks, with window sizes adaptively determined based on the volume resolution at each stage.  $\operatorname{SA}$ is followed by 3D convolutional blocks to generate the occupancy probability:
$P_{V} = \mathcal{CB}_{3D}(\operatorname{SA}(F_V^{\text{init}}))$,
which can be binarized to get the predicted voxel $\tilde{V}$.
This local spectrum-attention mechanism efficiently suppresses high-frequency noise while preserving textural details, addressing the inherent coexistence of texture details and high-frequency noise within the spectral domain of each LiDAR scan.

\vspace{1mm}\noindent\textbf{Flexible Resolution Uplifter.}
The quantization error is inevitably introduced, where $\|x_i-\tilde{x}_i\|\leq\frac{vs}{2}$, with $vs$ representing the voxel size and $x_i$ being the voxel center location. To improve reconstruction fidelity, we employ a lightweight upsampler that performs point upsampling and per-point offset prediction, formulated as: 
    $U: v_i \mapsto \{v_i+\Delta_{i,j}\}_{j=1}^f$.
Here, $v_i$ denotes the location of non-occupied reconstructed voxels and $f$ denotes the upsampling rate. This not only yields improved accuracy but also offers application-specific flexibility by allowing customizable point density. 

\subsection{Loss Function}\label{sec:loss}
We utilize the standard Lagrangian multiplier-based Rate-Distortion loss to obtain a trade-off between reconstruction quality and compression ratio:
    $L=D+\lambda{R}$,
where $D$ denotes distortion term, and $R$ represents bitrate penalty.

\vspace{1mm}\noindent\textbf{Rate Loss.} The total rate is calculated by adding the rates of the global features and the high-frequency details. To ensure the quantization step remains differentiable, we approximate it with additive uniform noise in the range $\lbrack{-0.5}, 0.5\rbrack$. The bitrates for feature maps are lower-bounded by the entropy $R=E(-\log_2{P_X(X))}$. We can assign bitrates to global content features and high-frequency features accordingly to preserve information while keeping storage requirements minimal:
$R=\lambda_1\cdot{R_{X_{S}}}+\lambda_2\cdot{R_{X_S^{H}}}$.

\vspace{1mm}\noindent\textbf{Distortion Loss.} To handle the severe class imbalance caused by LiDAR scene sparsity, we use focal loss to concentrate training on surface boundaries:

    $D=\text{FL}(p_t) = -\alpha(1 - p_t)^\gamma\log(p_t)$,
where $p_t = p$ for ground-truth occupied voxels and $p_t = 1-p$ for non-occupied voxels, and $p$ is the predicted occupancy probability.

\section{Experimental Results}
\begin{figure*}[t]
    \centering
    \includegraphics[width=\linewidth, trim=0cm 0cm 0cm 0cm]{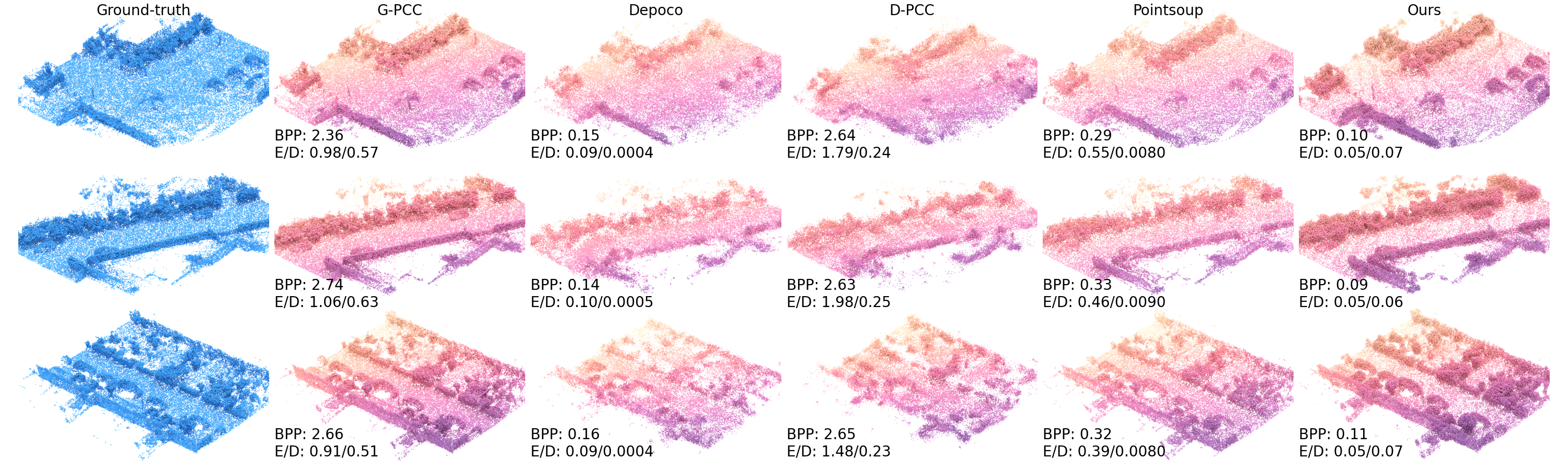}
    \vspace{-6mm}
    \caption{Qualitative results on the 40mILEN dataset. E/D denotes encoding and decoding time (in secs). Zoom in for details.}
    \vspace{-4mm}
    \label{fig:vis-40m}
\end{figure*}

\vspace{1mm}\noindent\textbf{Datasets.} We evaluate our method on three large-scale outdoor scene datasets using their official training and testing splits: the KITTI\_40mILEN (40mILEN)~\cite{wiesmann2021deep}, the SemanticKITTI (SemKITTI)~\cite{behley2019semantickitti}, and the Ford~\cite{pandey2011ford} dataset (See detailed configurations in supplementary).

\vspace{1mm}\noindent\textbf{Baselines.}
Our comparative analysis includes traditional TMC13~\cite{mpeg-pcc-tmc13} from MPEG PCC Group as the anchor.

For learning-based approaches, we compare our method against either results reported in the original papers or those obtained by retraining the models ourselves, including RENO~\cite{you2025reno}, Pointsoup~\cite{you2024pointsoup}, EHEM~\cite{song2023efficient}, SparsePCGC~\cite{wang2022sparse}, OctAttention~\cite{fu2022octattention}, D-PCC~\cite{he2022density}, and Depoco~\cite{wiesmann2021deep}.

To ensure a fair comparison in terms of runtime, we benchmarked all methods on the same RTX 4090 device, unless otherwise specified.

\vspace{1mm}\noindent\textbf{Test Conditions.}
To obtain anchor results, we implement the TMC13 configurations, with the angular coding mode disabled following~\cite{wang2022sparse}. We evaluate the rate-distortion performance using pairs of Bits-Per-Point (BPP) and PSNR. To assess the overall performance across the rate-distortion curve, we report BD-Rate (Rate) and BD-PSNR (PSNR) gains~\cite{bjontegaard2001calculation}.

\begin{table}[h]
\footnotesize
    \centering
    \caption{Runtime and Rate-Performance on 40mILEN.}
    \label{tab:kitti_40milen}
    \begin{tabular}{c|cccc}
    \toprule
        Methods  & Enc/Dec (s) & Rate Gain ($\downarrow$) & PSNR Gain ($\uparrow$) \\
    \midrule
    G-PCC        & 2.25/1.43    &   0.00\%        &      0.00     \\
    Depoco      & \underline{0.17}/\textbf{0.00086}    & +54.04\%    &      -2.06        \\
    D-PCC     &  1.26/0.24   &   +89.70\%     & -0.46          \\
    Pointsoup   & 2.58/\underline{0.0045} & \underline{-64.11\%} & \textbf{+3.40} \\
    \midrule
    Ours (w/o R)    & \textbf{0.065}/0.12 & \textbf{-94.21\%} & +1.40 \\  
     Ours & \textbf{0.065}/0.19 &  -93.73\% & \underline{+1.89}\\ 
    \bottomrule
    \end{tabular}

\end{table}
\subsection{Quantitative and Qualitative Results.}

\noindent \textbf{Rate-Distortion Performance.}

\begin{figure}[tbp]
\footnotesize
    \centering
    \includegraphics[width=\linewidth, trim=0cm 1cm 0 0]{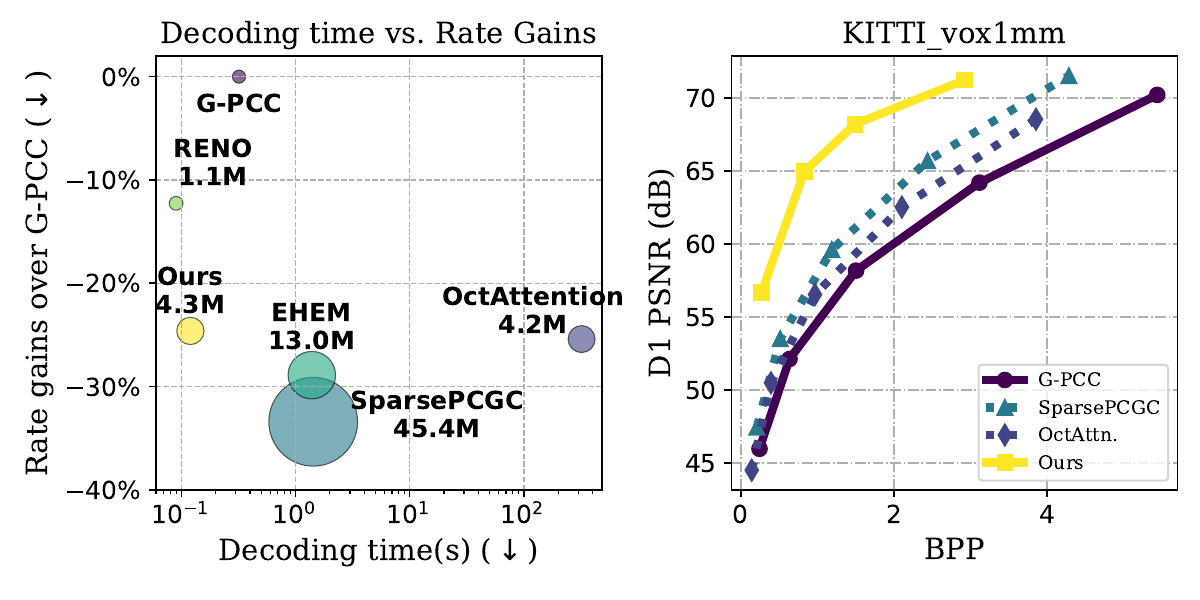}
    \caption{Left: Decoding efficiency and model footprint on SemanticKITTI, circle size indicates model size. Right: Rate-distortion curve on SemanticKITTI\_vox1mm.}
    \label{fig:semkitti-rd}
\end{figure}

Table~\ref{tab:kitti_40milen} and Figure~\ref{fig:semkitti-rd} report the overall rate gains on 40mILEN and SemKITTI, respectively. The results demonstrate that our method consistently outperforms all baseline approaches. Specifically, our method achieves a BD-rate improvement of 94.21\% over G-PCC on 40mILEN, and 78.51\% over JPEG on SemKITTI.

Figure~\ref{fig:semkitti-rd} (right) compares BPP–PSNR pairs with state-of-the-art methods, while Table~\ref{tab:kitti} compares against widely adopted LiDAR compression models summarized in a recent survey~\cite{wang2025compression}. To enable a fair comparison with voxel-based methods in Figure~\ref{fig:semkitti-rd}, we also report results for our model without the refinement technique (i.e., the point upsampler), denoted as Ours (w/o R). Similarly, VCN and SparsePCGC are evaluated without their Coordinates Refinement Module (CRM). The results also suggest that our approaches significantly outperform the baselines even without the refinement module. 

\begin{table}[tb]
\footnotesize
\addtolength{\tabcolsep}{0.5pt}
\centering
\caption{Comparison to mainstream methods on SemKITTI.} 
\label{tab:kitti}
\begin{tabular}{l|cc}
\toprule
Method & Type & Rate Gain (\%) \\
\midrule
JPEG XL~\cite{alakuijala2020benchmarking} & Image & 0.00\% \\
PCC-Cluster~\cite{sun2019novel} & Image & -47.44\% \\
R-PCC~\cite{wang2022r}  & Image & -47.35\% \\ 
Draco~\cite{Draco} & Tree & -40.84\% \\
PCC-SCSP~\cite{chen2022point}  & Tree  & -72.94\% \\
OctFormer~\cite{cui2023octformer} & Tree & -73.76\% \\
MLGE~\cite{fan2023multiscale}  & Tree & \underline{-75.21}\% \\ \midrule
Ours & Triplane &  \textbf{-78.51}\%\\  
\bottomrule
\end{tabular}
\end{table}

\noindent \textbf{Compression Latency.}
Tables~\ref{tab:kitti} and~\ref{tab:ford} demonstrate the low coding latency of our method. Owing to voxelization and triplane-based representation, the speed remains invariant to input point cloud resolution. 
Our method supports real-time coding and decoding, considering that LiDAR data is collected at a rate of 10 Hz.

\begin{figure}[tbp]
\footnotesize
    \centering
    \includegraphics[width=\linewidth]{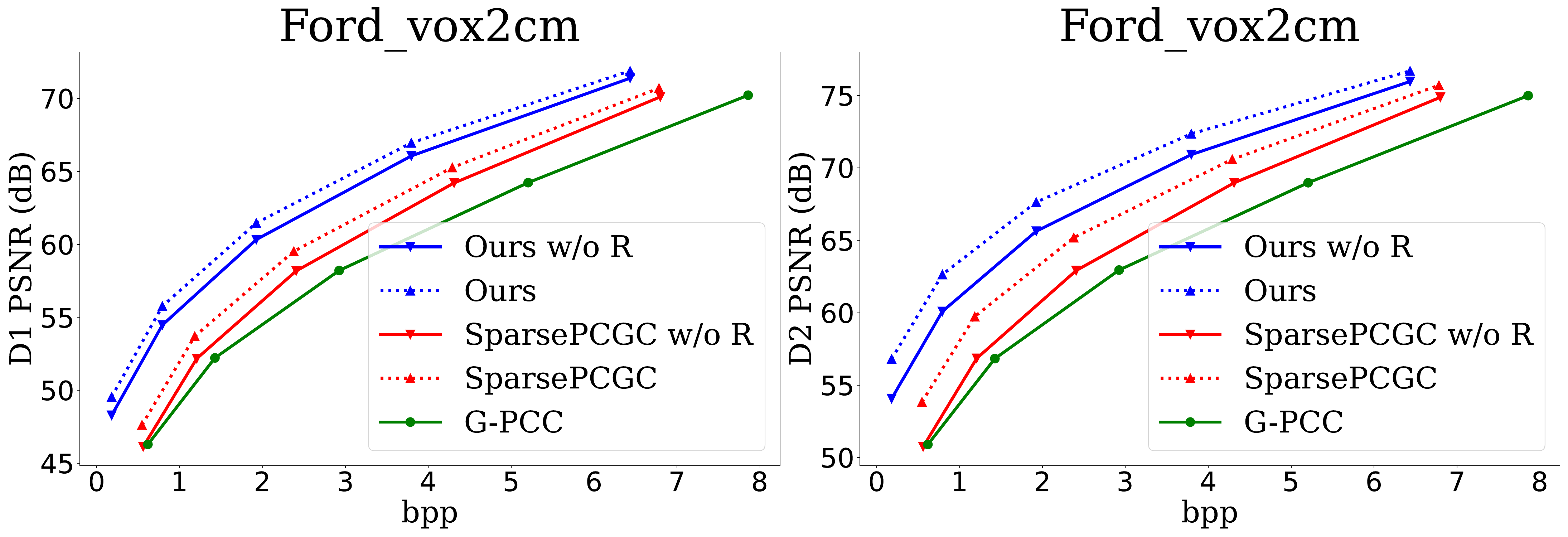}
    \vspace{-6mm}
    \caption{Generalization ability test on Ford Dataset.}
    \label{fig:ford-rd}
    \vspace{-4mm}
\end{figure}

\noindent \textbf{Generalization Abilities.}
Table~\ref{tab:ford} and Figure~\ref{fig:ford-rd} present the generalization results on the Ford dataset. Following SparsePCGC~\cite{wang2022sparse}, we evaluate on quantized Ford with 18-bit and 14-bit depth. Our method demonstrates robustness and generalizes well to unseen LiDAR scenes. With the proposed point upsampler, our method further improves the PSNR gains for different resolutions.

\vspace{1mm}\noindent\textbf{Qualitative Analysis.}
Figure~\ref{fig:vis-40m} presents the visualization results. We focus on visualizing 40mILEN, as it offers richer semantic content. The results indicate that our method effectively preserves street-level details such as vehicles, building boundaries, and roadside trees by accurately maintaining hierarchical frequency information. Additionally, our approach enables low bitrates and significantly reduced inference time. The triplane representation alleviates binary file size constraints, while the point upsampler allows reconstruction at arbitrary densities.

\begin{table}[tbp]
\footnotesize
    \centering
    \caption{Runtime and Compression Performance on Ford.}
    \label{tab:ford}
    \addtolength{\tabcolsep}{-2pt}
    \begin{tabular}{l|c|cc|cc}
    \toprule
    \multirow{2}{*}{Ford\_vox2cm} & Time (s) & \multicolumn{2}{c|}{Rate Gain (\%)} & \multicolumn{2}{c}{PSNR} Gain\\ \cmidrule{2-6}
     & Enc/Dec & D1 & D2 & D1 & D2 \\ \midrule
      G-PCC & 0.11/\textbf{0.04} & 0.00 & 0.00 & 0.00 & 0.00 \\
      SparsePCGC (w/o R) &  0.31/0.25 & -15.25 & -15.42 & 1.55 & 1.59 \\
      SparsePCGC & 0.34/0.27 & -27.47 & -35.28 & 2.97 & 4.00 \\
      Ours (w/o R) & \textbf{0.065}/\underline{0.12} & \underline{-52.48} & \underline{-55.77} & \underline{5.70} &  \underline{6.24} \\
      Ours & \textbf{0.065}/0.17 &  \textbf{-58.65} & \textbf{-65.88} & \textbf{6.77} & \textbf{8.14} \\
    \midrule\midrule
    \multirow{2}{*}{Ford\_vox1mm} & Time (s) & \multicolumn{2}{c|}{Rate Gain (\%)} & \multicolumn{2}{c}{PSNR} Gain\\ \cmidrule{2-6}
     & Enc/Dec & D1 & D2 & D1 & D2 \\
    \midrule
    G-PCC & 0.20/\textbf{0.11} & 0.00 & 0.00 & 0.00 & 0.00\\
    SparsePCGC (w/o R) &  1.67/1.14 & -12.18 & -12.30 & 1.62 & 1.64 \\
    SparsePCGC & 1.68/1.2 & -19.91 & -24.58  & 2.75 & 3.54  \\
    Ours (w/o R) & \textbf{0.065}/\underline{0.12} & \underline{-38.78} & \underline{-40.34} & \underline{4.96} & \underline{5.30} \\
    Ours & \underline{0.066}/0.17 & \textbf{-43.03} & \textbf{-46.80} & \textbf{5.78} & \textbf{6.76} \\
    \midrule
    \end{tabular}
\end{table}

\subsection{Ablation Study}
We assess the effectiveness of the proposed modules on KITTI\_40mILEN; all ablation variants are evaluated without the point upsampler, as it is an optional component already validated in Table~\ref{tab:kitti_40milen} and Table~\ref{tab:ford}. Intersection-over-Union (IoU) is reported following the grid settings in Depoco.

\vspace{1mm}

\begin{figure}
    \centering
    \includegraphics[width=\linewidth]{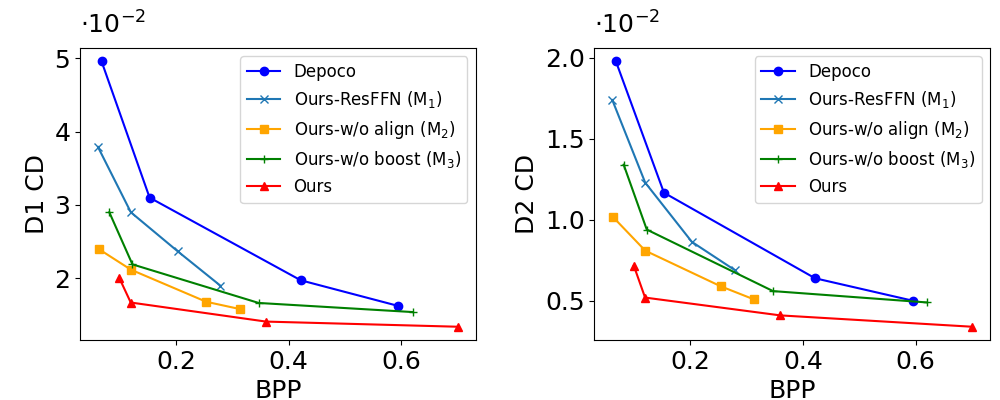}
    \vspace{-6mm}
    \caption{Effectiveness of boosting and alignment in FD and FM Block.}
    \vspace{-6mm}
    \label{fig:abl-freq}
\end{figure}

\noindent\textbf{Effectiveness of frequency decomposition and modulation block.} We conduct a systematic analysis of our frequency decomposition (FD) and frequency modulation (FM) blocks as shown in Figure~\ref{fig:abl-freq}. For variant 1 (M$_1$), we replace all the FD and FM blocks with standard residual convolution blocks. For variant 2 (M$_2$), we remove the frequency-boosted path in the FD Block while still decomposing planar features into high- and low-frequency components, which are subsequently integrated through the FM Block. 
For variant 3 (M$_3$), we turn off the high-frequency alignment in the FM block and instead directly include the high-frequency priors for feature reconstruction. 

\begin{figure}
    \centering
    \includegraphics[width=\linewidth]{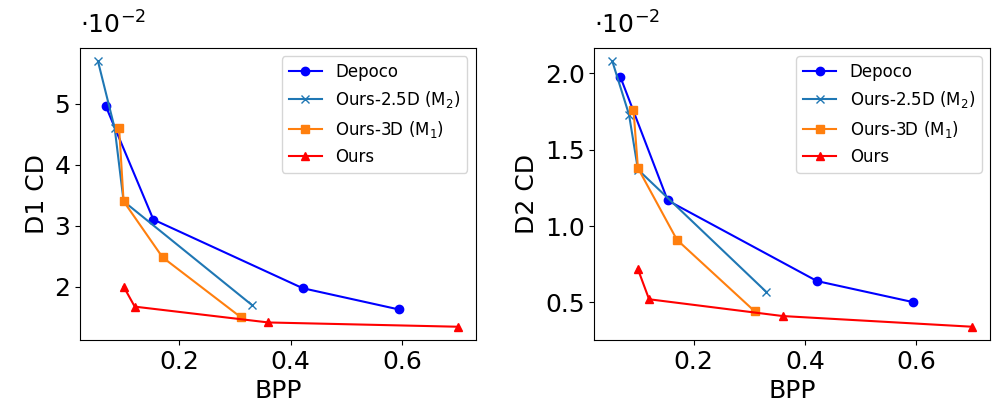}
        \vspace{-6mm}
    \caption{Effectiveness of triplane proxy.}
    \vspace{-2mm}
    \label{fig:abl-trip}
\end{figure}

\vspace{1mm}\noindent\textbf{Effectiveness of triplane representation.} We evaluate the impact of triplane representation on rate-distortion performance in Figure~\ref{fig:abl-trip}. For variant 1 (M$_1$), we constructed a purely 3D-based model by removing both the planar projector and back-projector components. Variant 2 (M$_2$) positions the planar projector before the entropy model and the back-projector after it, which we refer to as a 2.5D model. To ensure a fair comparison, we adjusted the number of basic channels to maintain similar parameter counts across all models. Results demonstrate that our approach of converting to triplane representations at the earliest stage achieves optimal rate-distortion performance. The fully 3D-based model requires substantially higher bitrates to maintain comparable reconstruction quality, primarily due to the tripling of total elements that need to be saved compared to the pixel-based approach. 

The 2D model reduces bitrate requirements compared to the 3D model, but still incurs significant computational overhead, necessitating a reduction in channel dimensions to maintain practicality. Both other approaches performed worse than early triplane conversion, highlighting that compression effectiveness depends more on channel richness than on the spatial resolution of the coded features.

\vspace{1mm}\noindent\textbf{Effectiveness of Parameter-Intensive Modules.} Table~\ref{tab:abl-paraml} shows the effectiveness of each parameter-intensive module. We are also interested in the importance of parameter-intensive modules and exploring the parameter trade-offs for different modules. 
We gradually remove the 3D LSA-Enhancer and position encoding (PE), which consume substantial parameters for high-resolution settings, and keep the total parameters unchanged by adjusting the basic feature dimensions. This simulates our decision process regarding modules when faced with limited memory. For low resolution, we use the full set to evaluate the importance of each part. Table~\ref{tab:abl-paraml} shows that 3D refinement remains the most important component in ensuring reconstruction quality, with basic dimensions and positional encoding being outlined next. We can easily conclude that the feature dimension matters more than the spatial dimension, likely because it directly influences the compression loss by affecting the local probability distribution of quantized features.

\begin{table}[tbp]
\footnotesize
\addtolength{\tabcolsep}{-2pt}
    \centering
    \caption{Effectiveness of Parameter-Intensive Modules.}
    \label{tab:abl-paraml}
    \begin{tabular}{c|ccc|c c}
    \toprule
         \multirow{4}{*}{\makecell{Resolution\\448x448x56}}&  LSAR Blocks  & PE & Basic Dim &  BPP & IoU ($\%$)  \\
         \cmidrule{2-6}
         & 2 & \checkmark & 32 & 0.061 & 39.7\\
         & 4 & & 24 & 0.060 & 43.2\\
         & 4 & \checkmark & 20  & \textbf{0.056} & \textbf{43.6}\\
    \midrule
         \multirow{4}{*}{\makecell{Resolution\\384x384x48}} &  LSAR Blocks  & PE & Basic Dim & BPP & IoU ($\%$) \\
         \cmidrule{2-6}
        & 3 & \checkmark & 40 & 0.065 & 48.5\\
         & 4 & & 36 & 0.063 & 49.1\\
         & 4 & \checkmark & 32  & \textbf{0.061} & \textbf{49.6}\\
     \midrule
            \multirow{4}{*}{\makecell{Resolution\\320x320x40}}  &  LSAR Blocks  & PE & Basic Dim & BPP & IoU ($\%$) \\
         \cmidrule{2-6}
         & 2 & \checkmark & 40 & 0.059 & 48.7\\
         & 4 & & 40 & 0053 & 49.4\\
         & 4 & \checkmark & 40  & \textbf{0.051} & \textbf{49.5}\\
    \bottomrule
    \end{tabular}
\end{table}

\section{Conclusion}
This study explores a frequency-disentangling framework for point cloud geometry compression, enabling flexible assignment of bitrate budgets and effective preservation of essential textures. By treating triplane features as proxies for LiDAR point cloud, our method achieves a 94\% reduction in bitrate and supports real-time compression at over 16 FPS. This approach signifies a notable advancement in decreasing bitrates and expediting the overall compression process. Building upon the frequency-aware planar autoencoder, we introduce a lightweight, frequency-based attention mechanism that enhances local 3D correlations and improves decompression quality. Our approach offers a novel and efficient prototype for point cloud compression, balancing adaptability, speed, and high-fidelity reconstruction.

\subsection{Acknowledgments}
This research was supported by the Australian Research Council (ARC) under discovery grant project \# 240101926. Professor Ajmal Mian is the recipient of an ARC Future Fellowship Award (project \# FT210100268) funded by the Australian Government.
\bibliographystyle{ieeetr}
\bibliography{ref}

\end{document}